\documentclass{article} %

\usepackage{microtype}
\usepackage{graphicx}
\usepackage{subcaption}
\usepackage{booktabs} 

\usepackage{hyperref}

\usepackage{amsmath}
\usepackage{amssymb}
\usepackage{mathtools}
\usepackage{amsthm}

\usepackage[capitalize,noabbrev]{cleveref}

\theoremstyle{plain}

\theoremstyle{definition}

\theoremstyle{remark}

\usepackage[textsize=tiny]{todonotes}

\usepackage{tabularx}
\usepackage{multirow}
\usepackage[most]{tcolorbox}
\usepackage[table]{xcolor}

\newcommand{\methodname}{START}

\usepackage{minitoc}

\usepackage{iclr2026_conference,times}
\usepackage{hyperref}
\usepackage{url}

\usepackage{amsmath,amsfonts,bm}

\def\eqref#1{equation~\ref{#1}}

\def\1{\bm{1}}

\DeclareMathAlphabet{\mathsfit}{\encodingdefault}{\sfdefault}{m}{sl}
\SetMathAlphabet{\mathsfit}{bold}{\encodingdefault}{\sfdefault}{bx}{n}

\renewcommand{\cite}{\ \citep}
\definecolor{mydarkred}{rgb}{0.6,0,0}
\definecolor{mydarkgreen}{rgb}{0,0.6,0}
\definecolor{mygray}{gray}{.9}

\hypersetup{
    colorlinks=true,
    linkcolor=mydarkred,
    citecolor=mydarkgreen
}

\title{RLVR Training of LLMs Does Not Improve Thinking Ability for General QA: Evaluation Method and a Simple Solution}

\author{
Kaiyuan Li\thanks{Equal contribution.}\,\,$^{1, 2}$ \quad 
Jing-Cheng Pang\footnotemark[1]\,\,$^{3}$ \quad 
Yang Yu\thanks{Corresponding author: \texttt{yuy@nju.edu.cn}}\,\,$^{1}$ \\
\\
$^1$National Key Laboratory for Novel Software Technology \& \\
\hphantom{$^1$}School of Artificial Intelligence, Nanjing University, China \\
$^2$Polixir.ai, China \\
$^3$Huawei Technologies Co., Ltd., China
}

\iclrfinalcopy %
\begin{document}

\doparttoc
\faketableofcontents

\hypersetup{linkcolor=black}
\maketitle
\hypersetup{linkcolor=mydarkred}

\begin{abstract}
Reinforcement learning from verifiable rewards (RLVR) stimulates the thinking processes of large language models (LLMs), substantially enhancing their reasoning abilities on verifiable tasks. It is often assumed that similar gains should transfer to general question answering (GQA), but this assumption has not been thoroughly validated. To assess whether RLVR automatically improves LLM performance on GQA, we propose a Cross-Generation evaluation framework that measures the quality of intermediate reasoning by feeding the generated thinking context into LLMs of varying capabilities. Our evaluation leads to a discouraging finding: the efficacy of the thinking process on GQA tasks is markedly lower than on verifiable tasks, suggesting that explicit training on GQA remains necessary in addition to training on verifiable tasks. We further observe that direct RL training on GQA is less effective than RLVR. Our hypothesis is that, whereas verifiable tasks demand robust logical chains to obtain high rewards, GQA tasks often admit shortcuts to high rewards without cultivating high-quality thinking. To avoid possible shortcuts, we introduce a simple method, Separated Thinking And Response Training (START), which first trains only the thinking process, using rewards defined on the final answer. We show that START improves both the quality of thinking and the final answer across several GQA benchmarks and RL algorithms.
\end{abstract}

\section{Introduction}
\label{sec:intro}

The paradigm of Large Language Models (LLMs)\cite{gpt4, llama3} has undergone a fundamental shift with the emergence of ``thinking'' models, which generate explicit internal thinking traces before producing a final response. Originally popularized by models like DeepSeek-R1\cite{r1} through Reinforcement Learning (RL)\cite{rl, star} on math and logic tasks, this approach utilizes clear, verifiable reward signals and advances the reasoning ability of LLMs to a new level. However, the effectiveness of this paradigm in general-purpose tasks—such as open-ended question answering or instruction following—remains inadequately understood.

To empirically validate whether these reasoning gains transfer to general domains, we propose a Cross-Generation Evaluation framework designed to isolate the intrinsic quality of the thinking process. This method operates by feeding different thinking traces generated by the source model into different models, treating the ``thought'' as a prompt for the ``responder''. The underlying premise is that high-quality reasoning should be universally beneficial, acting as a scaffold that improves the performance of any model. However, our evaluation reveals a discouraging disparity: while thinking traces learned via RLVR on verifiable tasks significantly boost performance in reasoning contexts, their efficacy drops precipitously when applied to GQA. Specifically, we observe that the marginal performance gains derived from employing a stronger thinking traces are significantly overshadowed by the improvements achieved through utilizing a more capable answering model. This finding challenges the assumption of automatic transfer and suggests that explicit training on GQA tasks remains essential alongside verifiable task training.

However, our experiments indicate that direct RL is markedly less effective than RLVR in fostering genuine thinking abilities. We identify a critical phenomenon termed \emph{thinking stagnation}: while direct RL yields substantial improvements in the quality of the final answers, the corresponding enhancement in the intermediate thinking process is disproportionately marginal. We hypothesize that this discrepancy arises from the weak coupling between thinking and response quality in general domains. Unlike verifiable tasks, where a correct solution relies strictly on a robust and error-free logical chain, GQA tasks often admit shortcuts to high rewards\cite{Unfaithfulness, RLHF_len}. In these scenarios, the model can learn to satisfy the reward function by optimizing the final answer only, bypassing the need to cultivate high-quality, structured thinking.

To mitigate the influence of these reward shortcuts, we propose Separated Thinking And Response Training (\methodname), a simple yet effective two-stage training paradigm. The core innovation of \methodname~lies in the decoupling of the thinking process from the response generation during the first training stage. Specifically, we first train only the thinking process while keeping the final answer generation fixed, using rewards derived from the resulting complete response. From an reinforcement learning perspective, this approach transforms the answer generation from a high-variance part of the model's exploration space into a stable component of the environment. Through this, we force the RL algorithm to attribute reward solely to the quality of the thinking trace, encouraging the model to cultivate genuine thinking capabilities robustly coupled with answer accuracy. Our empirical results demonstrate that \methodname~consistently outperforms joint RL baselines across a variety of RL algorithms and diverse GQA datasets, yielding both higher reward scores and superior answer quality. Furthermore, we observe that the more effective thinking traces cultivated by \methodname~exhibit a richer presence of meta-context\cite{reflexion, self_refine} suggesting that the model has learned to utilize its internal thought space for more sophisticated cognitive management.

The main contributions of this work are summarized as follows: (i) We propose a novel Cross-Generation evaluation framework to isolate the intrinsic value of thinking processes, revealing that unlike in verifiable tasks, the efficacy of the thinking process in GQA is significantly limited. (ii) We identify a ``thinking stagnation'' phenomenon where standard RL fails to effectively evolve the model's thinking process. (iii) We introduce \methodname, a two-stage training paradigm that decouples the thinking process from response generation by treating the final answer as a stable part of the environment instead of model exploration.


\section{The Thinking Stagnation of RL on GQA}
\label{sec:verification}

In this section, we present an empirical investigation into the role of internal thinking processes in GQA. We first introduce the \textbf{Cross-Generation} evaluation framework to quantify the actual contribution of thinking traces to final answer quality, revealing that the reasoning efficacy observed in verifiable tasks does not naturally translate to the GQA domain. Following this, we analyze the performance of direct reinforcement learning on GQA datasets. Our findings uncover a \textbf{Thinking Stagnation} phenomenon, where the model's thinking quality remains stagnant despite overall reward improvements.

\subsection{Uncovering the Limited Efficacy of Thinking in GQA}
\label{subsec:thinking_efficacy}

\begin{figure}[t]
    \centering
    \begin{subfigure}[b]{0.48\linewidth}
        \centering
        \includegraphics[width=\linewidth]{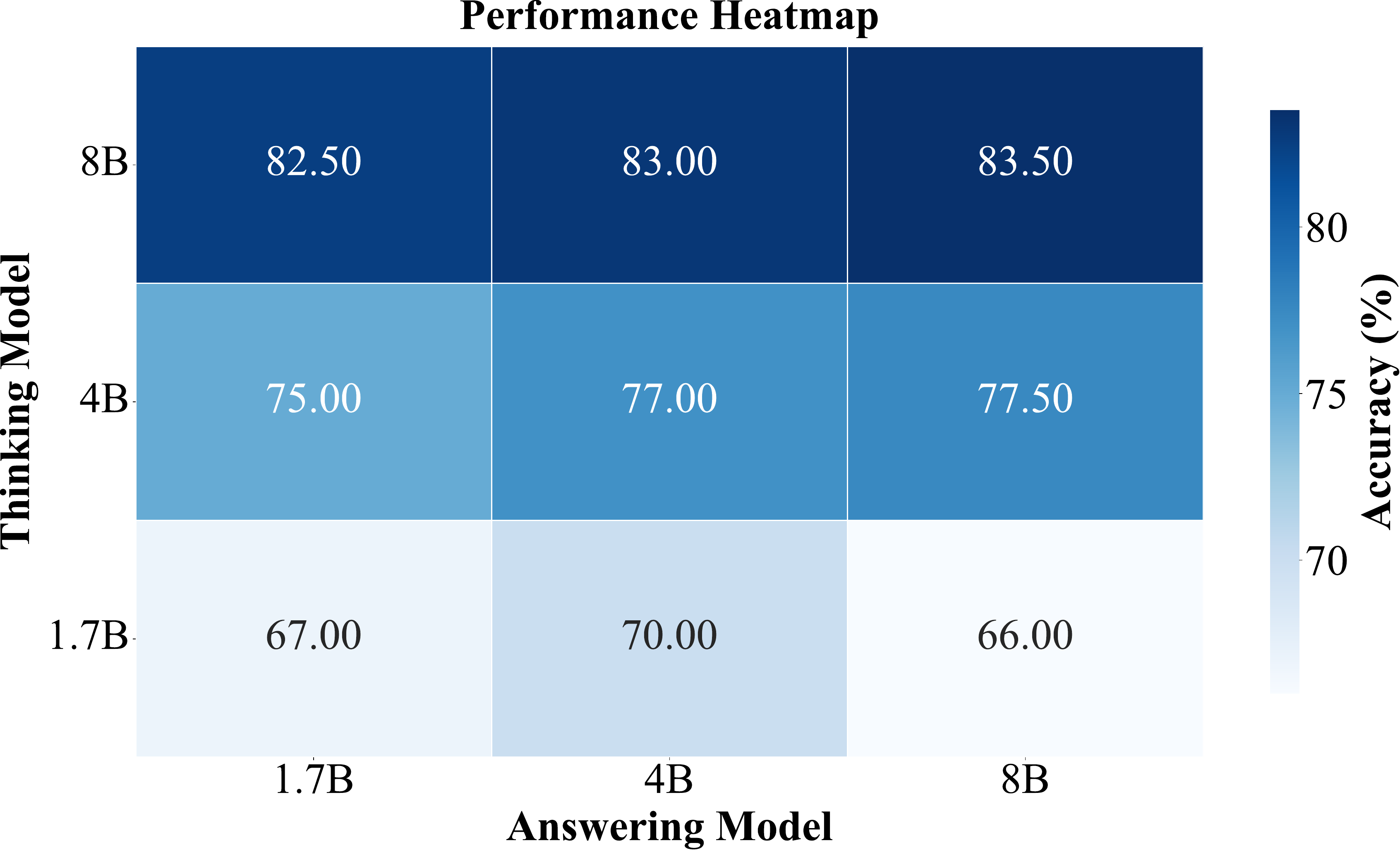}
        \caption{Reasoning (MATH)}
        \label{fig:heatmap_reason}
    \end{subfigure}
    \hfill
    \begin{subfigure}[b]{0.48\linewidth}
        \centering
        \includegraphics[width=\linewidth]{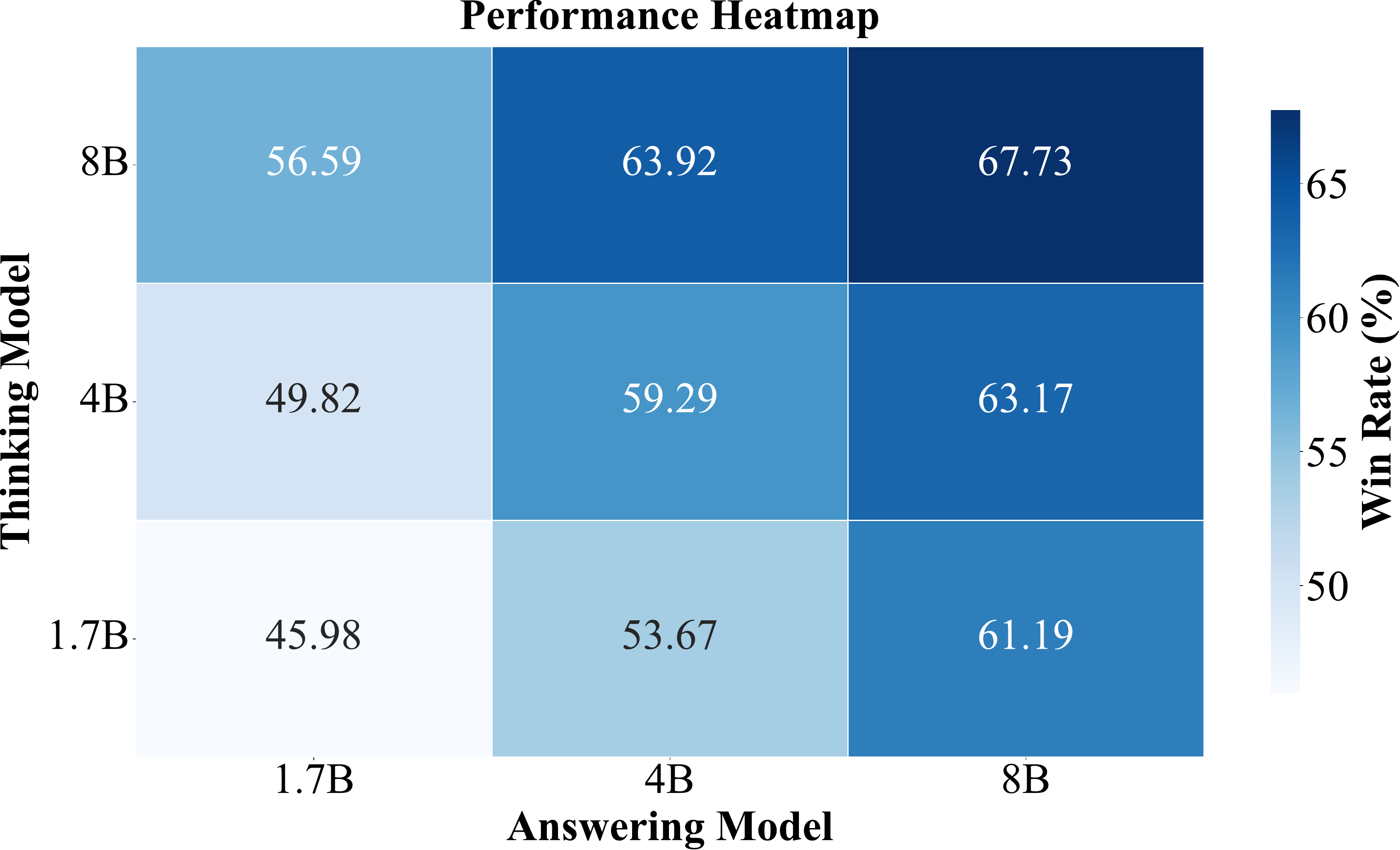}
        \caption{General QA (AlpacaEval 2.0)}
        \label{fig:heatmap_gqa}
    \end{subfigure}
    \caption{\textbf{Cross-generation performance heatmaps} Comparing the influence of thinking vs. answering model capacity. 
    (Left) In reasoning tasks, performance is almost entirely dictated by the quality of the thinking trace. 
    (Right) In general tasks, the answering model capacity remains a dominant factor, and the performance provided by superior thinking is less decisive.}
    \label{fig:heatmaps}
\end{figure}

In this subsection, we investigate whether the significant reasoning enhancements observed in verifiable tasks through RLVR naturally generalize to the broader domain of General Question Answering (GQA).

To empirically disentangle the quality of thinking from the final response, we implement a ``Cross-Generation'' evaluation framework. Specifically, for a given answer sequence $T_{X} + A_{Y}$, $T$ and $A$ represent the thinking trace and the final answer respectively. We first allow model $X$ to generate the thinking trace, then this trace is provided as a fixed prefix (pre-filled context) to model $Y$, which is tasked to generate the final answer.

To empirically isolate the cognitive utility of thinking traces, we utilize the Qwen3 model series (1.7B, 4B, and 8B) \cite{qwen3}. These models, by virtue of their varying parameter scales, naturally generate thinking traces of distinct quality and depth. For reasoning tasks, we evaluate performance on the MATH dataset\cite{math} using answer accuracy as the standard metric. For GQA, we employ the AlpacaEval 2.0 benchmark\cite{alpaca_eval}, reporting the standard length-controlled win rate against \textit{GPT-4-1106-preview}\cite{gpt4} to ensure a robust measure of general answering quality. Detailed hyperparameters and experimental configurations are provided in Appendix \ref{app:data}. By implementing the "Cross-Generation" framework, we construct performance heatmaps that visualize the interaction between thinking and answering efficacy across both domains.

The results from our experiments (Figure \ref{fig:heatmaps}) reveal a fundamental disparity in how thinking traces translate to final output quality across different domains. In the reasoning tasks (MATH), a high-quality thinking trace serves as the primary determinant of success. Upgrading the thinking trace from a 1.7B-level to an 8B-level for a 1.7B answering model results in a massive performance surge, from 67.00\% to 82.50\%. Crucially, once a high-quality thinking trace is provided, increasing the answering model's capacity yields almost no additional benefit—the accuracy only marginally improves from 82.50\% (1.7B model) to 83.50\% (8B model). This indicates that in formal reasoning, the thinking process is the clear ``bottleneck'', and its potential is already being effectively exploited. However, the dynamics in the GQA (AlpacaEval) are strikingly different. While a stronger thinking trace does improve the win rate (e.g., from 45.98\% to 56.59\% for a 1.7B answering model), this improvement is far less decisive. In fact, the gain from upgrading the answering model while keeping the thinking trace fixed is significantly more pronounced: the same 8B thinking trace paired with an 8B answering model jumps to 67.73\%, a much larger leap than the gain provided by the thinking trace alone.

These observations suggest that in the general domain, the ``thinking-to-answering'' transition is highly inefficient. The latent potential and efficacy of thinking traces in GQA remain largely untapped compared to their impact in reasoning tasks. This underscores the necessity for additional, targeted reinforcement learning training in GQA to bridge this cognitive gap and ensure the model to leverage its internal thinking more effectively.

\subsection{The Stagnation of Thinking in Reinforcement Learning on GQA}
\label{subsec:rl_stagnation}

Building upon the findings in Section \ref{subsec:thinking_efficacy}, which revealed the limited efficacy of existing thinking processes in the GQA domain, we conduct further reinforcement learning experiments directly on GQA datasets, aiming to determine if targeted RL signals can force the model to evolve more effective and substantive thinking traces that lead to better task performance.

We continue to use the ``Cross-Generation'' evaluation framework specified in section \ref{subsec:thinking_efficacy}. This protocol allows us to verify whether an RL-tuned model has evolved a superior ``thinking engine'', or if the improvements are merely localized to the answering phase. To quantify performance, we report both the reward achieved after training and the Win Rate, defined as the frequency with which a specific configuration's output is preferred over the output of the Base model in a pairwise comparison.

\begin{table}[ht]
    \centering
    \small
    \caption{\textbf{Stagnation of thinking evolution in general tasks.} Experimental verification of the thinking-answering decoupling effect. We report Reward and Win Rate (vs. Base) across various model scales, RL algorithms, and datasets. $\Delta$ denotes the reward improvement relative to the Base model. Subscripts denote the model state before (\textit{pre}) and after (\textit{post}) RL fine-tuning.}
    \label{tab:rl_think_fail}
    \begin{tabularx}{0.5\columnwidth}{l|ccc}
    \toprule
    \textbf{Configuration} & \textbf{Reward} & \textbf{$\Delta$ R} & \textbf{Win Rate} \\
    \midrule
    \multicolumn{4}{l}{\textit{Setting A: Qwen3-1.7B / GRPO / ExpertQA}} \\
    Base & 0.1842 & --- & --- \\
    $T_{\text{post}}$ + $A_{\text{pre}}$ & 0.1849 & \textbf{+0.0007} & 52.87\% \\
    $T_{\text{pre}}$ + $A_{\text{post}}$ & 0.2148 & +0.0306 & 100.0\% \\
    Post-trained & 0.2182 & +0.0340 & 100.0\% \\
    \midrule
    \multicolumn{4}{l}{\textit{Setting B: Qwen3-1.7B / GRPO / UltraFeedback}} \\
    Base & 0.1535 & --- & --- \\
    $T_{\text{post}}$ + $A_{\text{pre}}$ & 0.1539 & \textbf{+0.0004} & 49.25\% \\
    $T_{\text{pre}}$ + $A_{\text{post}}$ & 0.1677 & +0.0142 & 86.50\% \\
    Post-trained & 0.1658 & +0.0123 & 76.25\% \\
    \midrule
    \multicolumn{4}{l}{\textit{Setting C: Qwen3-1.7B / DAPO / ExpertQA}} \\
    Base & 0.1842 & --- & --- \\
    $T_{\text{post}}$ + $A_{\text{pre}}$ & 0.1868 & \textbf{+0.0026} & 58.92\% \\
    $T_{\text{pre}}$ + $A_{\text{post}}$ & 0.2261 & +0.0419 & 100.0\% \\
    Post-trained & 0.2352 & +0.0510 & 100.0\% \\
    \midrule
    \multicolumn{4}{l}{\textit{Setting D: Qwen3-4B / GRPO / ExpertQA}} \\
    Base & 0.1924 & --- & --- \\
    $T_{\text{post}}$ + $A_{\text{pre}}$ & 0.1921 & \textbf{-0.0003} & 47.77\% \\
    $T_{\text{pre}}$ + $A_{\text{post}}$ & 0.2213 & +0.0289 & 99.36\% \\
    Post-trained & 0.2224 & +0.0300 & 96.50\% \\
    \bottomrule
    \end{tabularx}
\end{table}

As presented in Table \ref{tab:rl_think_fail}, the results on general task datasets (ExpertQA\cite{expertqa} and UltraFeedback\cite{ultrafeedback}) reveal a striking phenomenon of thinking-answering decoupling. Across various model scales (Qwen3-1.7B and 4B) and RL objectives (GRPO and DAPO\cite{dapo}), the configuration $T_{\text{post}} + A_{\text{pre}}$ yields nearly negligible gains in Reward and Win Rate. For instance, in Setting A (Qwen3-1.7B on ExpertQA), the Reward only marginally increases from 0.1842 to 0.1849 ($+0.0007$). This suggests that the thinking traces generated by the post-trained model offer no more cognitive utility than those from the base model. Conversely, the $T_{\text{pre}} + A_{\text{post}}$ configuration captures the vast majority of the post-training gains, often matching or even exceeding the performance of the full post-trained model.

\begin{table}[ht]
    \centering
    \small
    \caption{\textbf{Comparison of thinking evolution across different models and domains.} \textbf{$\Delta R\%$} denotes the percentage change in reward relative to the \textit{Base} configuration.}
    \label{tab:rl_think_succ}
    \begin{tabularx}{0.52\columnwidth}{l|ccc}
    \toprule
    \textbf{Configuration} & \textbf{Reward} & \textbf{$\Delta$ R\%} & \textbf{Win Rate} \\
    \midrule
    \multicolumn{4}{l}{\textit{Setting A: Hunyuan-1.8B-Instruct / ExpertQA}} \\
    Base & 0.1628 & --- & --- \\
    $T_{\text{post}}$ + $A_{\text{pre}}$ & 0.1740 & \textbf{+6.88\%} & 69.11\% \\
    $T_{\text{pre}}$ + $A_{\text{post}}$ & 0.1942 & +19.29\% & 93.31\% \\
    Post-trained & 0.2111 & +29.67\% & 98.41\% \\
    \midrule
    \multicolumn{4}{l}{\textit{Setting B: DeepSeek-R1-Distill-Qwen-1.5B / ExpertQA}} \\
    Base & 0.1327 & --- & --- \\
    $T_{\text{post}}$ + $A_{\text{pre}}$ & 0.1478 & \textbf{+11.38\%} & 71.97\% \\
    $T_{\text{pre}}$ + $A_{\text{post}}$ & 0.1536 & +15.75\% & 90.13\% \\
    Post-trained & 0.1664 & +25.40\% & 92.68\% \\
    \toprule
    \textbf{Configuration} & & \textbf{$\Delta$ Acc} & \textbf{Accuracy} \\
    \midrule
    \multicolumn{4}{l}{\textit{Setting C: Qwen3-1.7B / MATH}} \\
    Base & --- & --- & 55.00\% \\
    $T_{\text{post}}$ + $A_{\text{pre}}$ & --- & \textbf{+7.60} & 62.60\% \\
    $T_{\text{pre}}$ + $A_{\text{post}}$ & --- & +7.45 & 62.45\% \\
    Post-trained & --- & +9.20 & 64.20\% \\
    \bottomrule
    \end{tabularx}
\end{table}

As shown in Table \ref{tab:rl_think_succ}, we extend our analysis to models with different architectural backgrounds, such as Hunyuan-1.8B-Instruct\cite{hunyuan} and DeepSeek-R1-Distill-Qwen-1.5B\cite{r1}. While these models show slightly more ``active'' thinking traces compared to Qwen3, with the $T_{\text{post}} + A_{\text{pre}}$ configuration yielding relative reward gains of +6.88\% and +11.38\% respectively, these improvements are still significantly overshadowed by the gains in the answering phase. In Setting A, for instance, the $T_{\text{post}} + A_{\text{pre}}$ configuration achieves a +19.29\% reward increase, almost triple what the evolved thinking alone contributes. This indicates that even when some degree of thinking evolution occurs, the RL process continues to prioritize the answering part as the primary vehicle for reward acquisition. In sharp contrast, for reasoning-intensive tasks like Setting C (MATH), the thinking process acts as the decisive engine for performance: the $T_{\text{post}} + A_{\text{pre}}$ configuration accounts for the vast majority of the total gain. In this domain, the model is effectively forced to evolve its thinking process to achieve success.

This empirical evidence suggests a clear \textbf{Thinking Stagnation} where the RL process primarily optimizes the model's ability to generate high-reward answers from existing thinking traces, rather than evolving the thinking process itself to provide better logical foundations.
\section{Separated Thinking And Response Training (\methodname)}
\label{sec:method}

\begin{figure}[t]
	\centering
	\includegraphics[width=\linewidth]{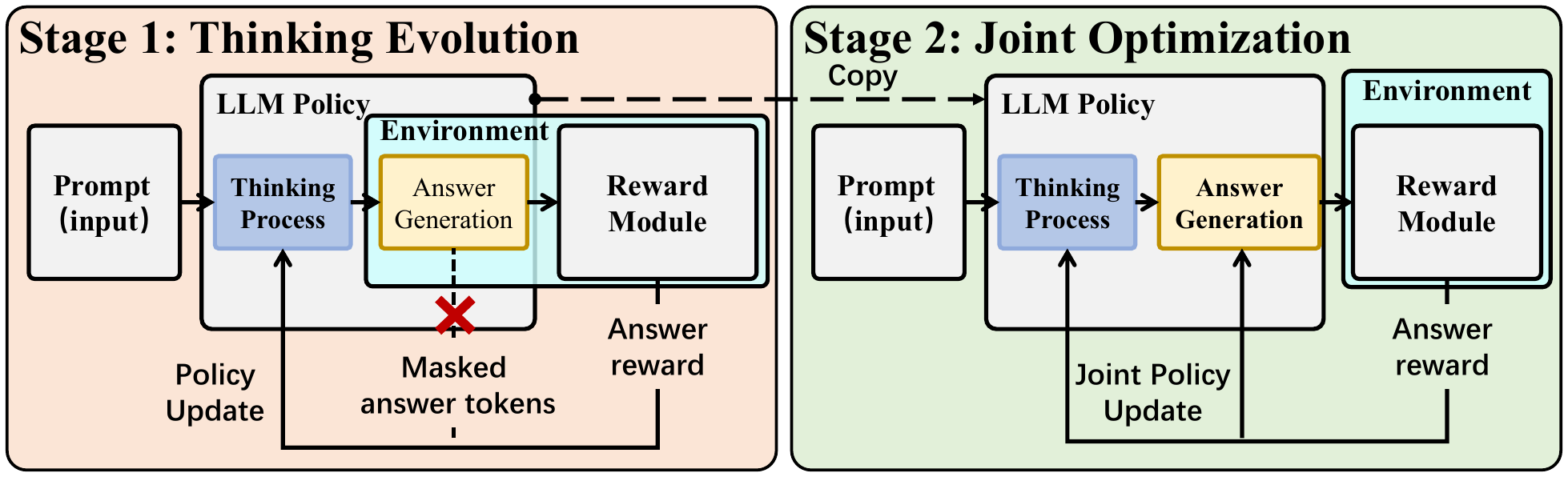}
	\caption{Overall framework of \methodname~method.}
 \label{fig:overall_framework}
\end{figure}

Based on the Thinking Stagnation uncovered in Section \ref{sec:verification}, we introduce \methodname~(\underline{S}eparated \underline{T}hinking \underline{A}nd \underline{R}esponse \underline{T}raining). This framework decouples the training of cognitive thinking from response generation through a two-phase optimization process.

\subsection{Preliminaries: RL for LLMs as an MDP}
\label{subsec:preliminary}

To formalize the proposed method, we take an reinforcement learning perspective of LLMs and formalize the language generation task as an Markov Decision Process (MDP)\cite{MDP, mdp_llm}, defined by the tuple $(\mathcal{S}, \mathcal{A}, P, R, \gamma)$:
\begin{itemize}
    \item \textbf{State Space $\mathcal{S}$:} A state $s_t$ at step $t$ consists of the initial instruction $x$ and the tokens generated so far, i.e., $s_t = [x, y_1, \dots, y_{t-1}]$.
    \item \textbf{Action Space $\mathcal{A}$:} The action corresponds to the vocabulary $\mathcal{V}$ of the LLM. At each step $t$, the model samples a token $y_t \in \mathcal{V}$.
    \item \textbf{Transition Dynamics $P$:} In language generation, the transition is deterministic where the next state $s_{t+1}$ is formed by appending the action $y_t$ to the current sequence.
    \item \textbf{Reward Function $R$:} We focus on outcome-based RL where a scalar reward $r(x, s)$ is provided only after the complete sequence $s = [T, A]$ is generated, where $T$ denotes thinking tokens and $A$ denotes answer tokens.
    \item \textbf{Policy $\pi_\theta$:} The policy $\pi_\theta(y_t | s_t)$ represents the LLM being optimized, which defines the probability distribution over the vocabulary given the current state.
\end{itemize}

In standard end-to-end RL frameworks, the goal is to maximize the expected reward $\mathbb{E}_{s \sim \pi_\theta} [r(x, s)]$. Under this formulation, the entire sequence $s = [T, A]$ is treated as the policy's actions, meaning both thinking and answering reside within the same \textbf{exploration space}.

\subsection{Conceptual Shift: From Exploration Space to Environment}
\label{subsec:env_shift}

While in verifiable tasks, a correct solution relies strictly on a robust and error-free logical chain, the weak coupling between thinking and answering in general tasks implies that the model is not strictly required to explore superior thinking processes to achieve higher rewards. Consequently, the reward signal becomes more directly associated with the answer tokens, allowing the model to improve performance while the thinking process remains in a state of stagnation.

To break this shortcut, \methodname~proposes a fundamental shift in the optimization perspective: \textbf{transforming the answering phase from a part of the policy's exploration space into a fixed component of the environment}. During the initial stage of training, the policy $\pi_\theta$ is only considered responsible for the ``action'' of generating the thinking trace $T$. The subsequent generation of the answer $A$ is no longer treated as part of model exploration, but is instead redefined as part of the environmental dynamics.

From the perspective of the RL agent, the answering phase becomes a stationary transition process that maps a given thought $T$ to a final outcome. By reclassifying the answering phase as part of the environment, we effectively remove it from the reward optimization, making exploring and evolving superior thinking traces $T$ the only viable path for the policy to achieve higher rewards. This shift ensures that the optimization pressure is purely concentrated on the evolution of internal thinking.

\subsection{Phase I: Thinking Evolution via Gradient Masking}

To operationalize the conceptual shift described in Section \ref{subsec:env_shift}, we introduce a gradient masking mechanism during the first phase of training. Given an instruction $x$, the model generates a sequence $s = [T, A]$, where $T$ represents the thinking tokens and $A$ represents the final answer tokens. During the backward pass of the RL objective (e.g., GRPO), we apply a gradient mask to the answer part. Specifically, the loss function for Phase I is modified as:

$$\mathcal{L}_{\methodname} = \mathbb{E} \left[ \sum_{i \in T \cup A} M_i \cdot \nabla_{\theta} \log \pi_{\theta}(s_i | x, s_{<i}) \cdot \hat{A} \right]$$where $M_i$ is a binary mask defined as:$$M_i = \begin{cases} 1 & \text{if } s_i \in T \\ 0 & \text{if } s_i \in A \end{cases}$$

Through this masking, the answer $A$—while still sampled from the model—functions as a fixed ``response head'' that maps the generated thought $T$ to a reward. Because the gradients for tokens in $A$ are zeroed out, the model is physically unable to improve its reward by merely adjusting its answering. In this configuration, the only way for the policy to improve the advantage $\hat{A}$ is to evolve thinking traces that are more logically rigorous or informative. This phase ensures that the evolution of internal thinking is the sole driver of performance gains.

\subsection{Phase II: Joint Optimization}

Once the model has evolved a robust ``thinking engine'' in Phase I, we transition to Phase II: a standard full-parameter RL fine-tuning. In this stage, the gradient mask is removed ($M_i=1$ for all $i \in T \cup A$), allowing the model to jointly optimize both thinking and answering. The primary goal of Phase II is to align the answering with the newly acquired thinking capabilities, ensuring that the model can effectively extract and present the logical insights generated in $T$.

\subsection{Integrated with Standard RL Algorithms}

\methodname~is a simple but effective method, which could be naturally integrated with any existing RL algorithms for LLM optimization. Unlike Process-based Reward Models (PRMs) that require expensive step-level annotations, \methodname~utilizes existing outcome-based Reward Models. It requires only a minor modification to the loss calculation (a simple token-level mask), making it compatible with most modern RL frameworks like DAPO or other open-source GRPO implementations. This makes \methodname~an out-of-the-box solution for researchers seeking to boost thinking capabilities in general domains without complex engineering.

\section{Experiment}
\label{sec:exp}

\subsection{Setup}

We utilize Qwen3-1.7B as the seed model for all training iterations. To provide reliable and scalable feedback, we employ ArmoRM\cite{ArmoRM}, an 8B-parameter multi-reward model that outputs a scalar score for a single response, as our reward model. We experiment with GRPO and DAPO using ExpertQA and UltraFeedback, focusing on general question answering and instruction following with actual human instructions.

To further benchmark the efficacy of our method against specialized reasoning-alignment techniques, we include GRPO-MA \cite{grpo_ma} as a baseline. GRPO-MA is designed to address gradient coupling and unstable advantage estimation in standard GRPO by employing a multi-answer generation strategy. Specifically, for each generated thinking trace, the model samples multiple independent answers. The advantage of a ``thought'' is then calculated based on the aggregated performance of its associated answers, decoupling the quality of the reasoning process from the stochasticity of a single response.

For each configuration, the baseline consists of the same seed model trained under the identical RL algorithm but without the two-phase separation. All baselines are trained until convergence to ensure a fair comparison. More details about datasets, models, algorithms are provided in the appendix\ref{app:data}.

\subsection{Main Result}

\begin{figure}[ht]
    \centering
    \includegraphics[width=0.7\linewidth]{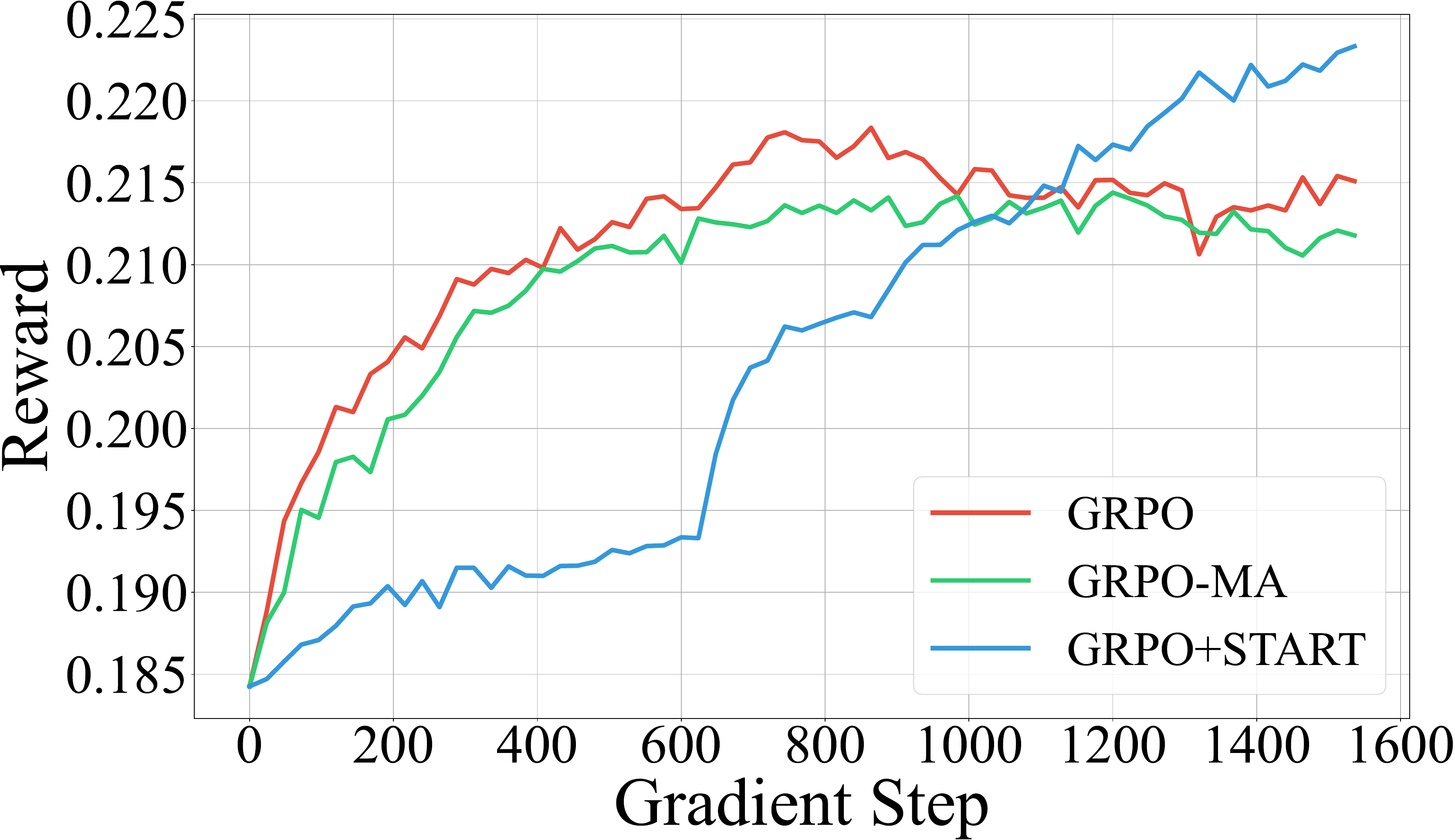}
    \caption{\textbf{Reward curves during reinforcement learning fine-tuning.} We compare the reward curves of GRPO, GRPO-MA, and our proposed \textbf{GRPO+\methodname} on the ExpertQA dataset.}
    \label{fig:reward_curve}
\end{figure}

The primary experimental results are summarized in Table \ref{tab:rl_main} and the reward convergence is visualized in Figure \ref{fig:reward_curve}. We analyze the performance of \methodname~across two critical dimensions: holistic performance and standalone thinking utility.

\begin{table}[h]
    \centering
    \small
    \caption{\textbf{Main results of \methodname~on ExpertQA.} Win Rates are calculated head-to-head against the post-trained GRPO baseline.}
    \label{tab:rl_main}
    \begin{tabularx}{0.61\columnwidth}{l|cc|cc}
    \toprule
    \multirow{2}{*}{\textbf{Configuration}} & \multicolumn{2}{c}{\textbf{Post-trained}} & \multicolumn{2}{c}{\textbf{$T_{\text{post}}$ + $A_{\text{pre}}$}} \\
    & Reward & Win Rate & Reward & Win Rate \\
    \midrule
    GRPO & 0.2182 & --- & 0.1849 & --- \\
    GRPO-MA & 0.2140 & 34.08\% & 0.1859 & 54.14\% \\
    \midrule
    GRPO+\methodname & \textbf{0.2201} & \textbf{59.24\%} & \textbf{0.1876} & \textbf{68.15\%} \\
    \bottomrule
    \end{tabularx}
\end{table}

The most significant finding is that \methodname~successfully overcomes the thinking stagnation identified in Section \ref{sec:verification}. As shown in the $T_{\text{post}} + A_{\text{pre}}$ column of Table \ref{tab:rl_main}, when pairing the post-trained thinking traces with a frozen base answering head, GRPO+\methodname~achieves a remarkable $68.15\%$ win rate against the vanilla GRPO baseline. These results provide direct empirical evidence that by masking the answering gradients in Phase I, \methodname~effectively forces the model to allocate its optimization capacity toward evolving its internal reasoning logic.

Our results further demonstrate that cultivating a superior thinking engine leads to a significantly higher performance ceiling for the final model. In the holistic "Post-trained" evaluation, GRPO+\methodname~achieves a win rate of $59.24\%$ over vanilla GRPO and reaches a higher reward of approximately $0.220$. As illustrated in Figure \ref{fig:reward_curve}, the reward curve for GRPO+\methodname~(blue line) exhibits a steady ascent during the first 600 steps of Phase I and a subsequent surge in Phase II, rapidly surpassing both the GRPO and GRPO-MA baselines. By first establishing a more robust logical foundation in Phase I, the model can more effectively align its answering head in Phase II, resulting in a final output that is not only stylistically preferred but also cognitively deeper.

We also compare \methodname~against GRPO-MA. While GRPO-MA shows a moderate improvement in standalone thinking utility ($54.14\%$ win rate in the $T_{\text{post}} + A_{\text{pre}}$ setting), it fails to match the gains of \methodname~($68.15\%$) and eventually reaches a lower reward plateau in the final post-trained model (as shown in Figure \ref{fig:reward_curve}). While GRPO-MA samples multiple answers per thought to provide a more stable and accurate estimate of the thought's value, its optimization remains joint. In the context of GQA, where the coupling between thinking and results is inherently weak, the model still finds it ``cheaper'' to optimize the answer tokens to capture the stable reward signal rather than evolving the complex logic required for high-quality thinking. Also, for a fixed total sampling budget ($G = K \times M$), GRPO-MA needs to sample multiple answers per thought ($M$), thus reduces the number of unique thinking paths ($K$) explored and reduces the diversity of final answers. This compression of the exploration space may explains why GRPO-MA slightly underperform standard GRPO baseline in certain GQA benchmarks.

\begin{figure}[t]
    \centering
    \begin{subfigure}[b]{0.48\linewidth}
        \centering
        \includegraphics[width=\linewidth]{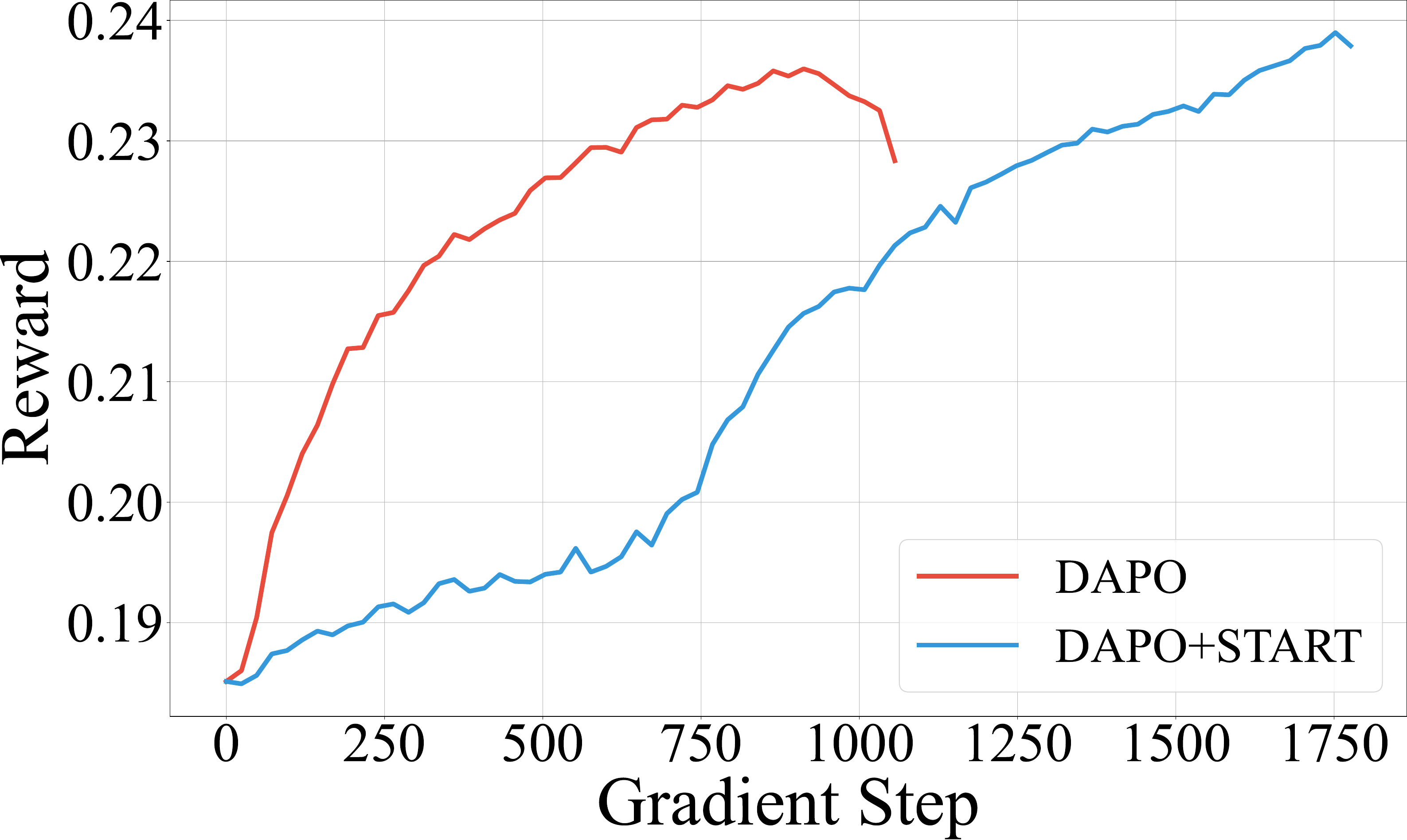}
        \caption{Reward curve using DAPO}
    \end{subfigure}
    \hfill
    \begin{subfigure}[b]{0.48\linewidth}
        \centering
        \includegraphics[width=\linewidth]{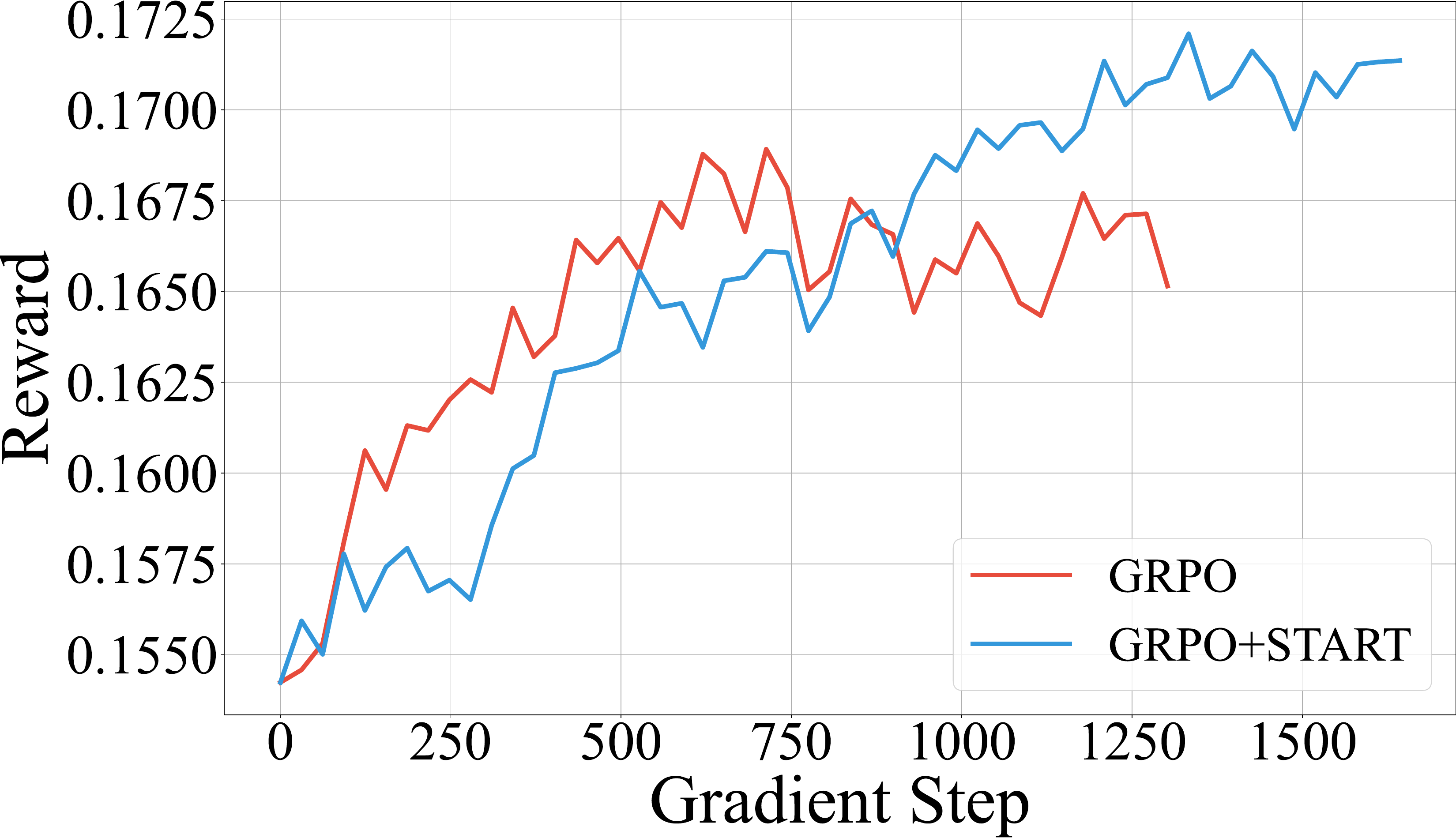}
        \caption{Reward curve on UltraFeedback}
    \end{subfigure}
    \caption{\textbf{Additional reward curves across different datasets and algorithms.} For baseline models, we terminate the training process if an obvious convergence trend is observed.}
    \label{fig:reward_curve_more}
\end{figure}

Beyond the primary GRPO results on ExpertQA, we extended our evaluation to include the DAPO algorithm and the UltraFeedback dataset to further test the versatility of our approach. These supplementary experiments consistently demonstrate that START yields significant performance gains across varying RL frameworks and data distributions, effectively validating its robustness and broad generalization capabilities in facilitating thinking evolution.

\begin{table}[ht]
    \centering
    \small
    \caption{\textbf{More results across different dataset and algorithm.} Win Rates are calculated head-to-head against the post-trained base algorithm.}
    \label{tab:rl_more}
    \begin{tabularx}{0.4\columnwidth}{l|cc}
    \toprule
    \textbf{Configuration} & \textbf{Reward} & \textbf{Win Rate} \\
    \midrule
    \multicolumn{3}{l}{\textit{ExpertQA / DAPO}} \\
    DAPO & 0.2352 & --- \\
    DAPO+\methodname & 0.2385 & 65.29\% \\
    \midrule
    \multicolumn{3}{l}{\textit{UltraFeedback / GRPO}} \\
    GRPO & 0.1658 & --- \\
    GRPO+\methodname & 0.1695 & 61.25\% \\
    \bottomrule
    \end{tabularx}
\end{table}

\subsection{Analysis of Phase I: Thinking as a Foundation}

\begin{table}[ht]
    \centering
    \small
    \caption{\textbf{Results of Phase I.} We decouple the contributions of thinking ($T$) and answering ($A$) by cross-pairing the base and post-trained (\methodname) components on ExpertQA using Qwen-1.7B. Win rates are evaluated relative to the base model. $\Delta R$ denotes the reward improvement.}
    \label{tab:rl_phase_I}
    \begin{tabularx}{0.51\columnwidth}{l|ccc}
    \toprule
    \textbf{Configuration} & \textbf{Reward} & \textbf{$\Delta$ R} & \textbf{Win Rate} \\
    \midrule
    Base & 0.1842 & --- & --- \\
    $T_{\text{post}}$ + $A_{\text{pre}}$ & 0.1888 & \textbf{+0.0046} & 69.11\% \\
    $T_{\text{pre}}$ + $A_{\text{post}}$ & 0.1897 & +0.0055 & 78.98\% \\
    Post-trained & 0.1927 & +0.0085 & 84.71\% \\
    \bottomrule
    \end{tabularx}
\end{table}

To further understand the dynamics of the \methodname~framework, we analyze the training characteristics and performance gains of Phase I (Thinking Evolution). in table \ref{tab:rl_phase_I}.

The results of the $T_{\text{post}} + A_{\text{pre}}$ configuration provide direct evidence of thinking evolution. By simply replacing the base model's thinking traces with those generated after Phase I training, the model achieves a significant reward increase of $+0.0046$ and a win rate of 69.11\%. This confirms that Phase I, through its gradient-masking mechanism, successfully forces the model to optimize its internal thinking independently of adjustments in the final answer. Importantly, we find that the thinking quality developed in Phase I merely degrade when the answering gradient is unmasked in Phase II. Instead, the improved thinking process serves as a robust foundation, allowing the answering head to extract more accurate insights and achieve a massive leap in performance. Moreover, as shown in our training curves in \ref{fig:reward_curve}, this masking strategy ensures a smooth and steady convergence of the thinking process, despite the loose coupling between thinking traces and final answers.

\subsection{Exploring the Emergence of Meta-Context Modeling}

To gain a more granular perspective on how thinking traces evolve under \methodname, we conduct a subsequent quantitative analysis. In general-purpose tasks, where logical rigor is often implicit, we explore whether the utility of thinking manifests through the discovery of Meta-Contexts—potential hidden variables such as user persona, intent, and structural preferences that, while not explicitly stated, may be essential for high-quality responses.

Specifically, we observe distinct patterns across the models. The Base Model often produces a passive "knowledge dump," while Vanilla GRPO exhibits a nascent awareness of style. Intriguingly, the START-trained model’s thinking traces appear to function as a strategic workspace where these hidden dimensions are more explicitly modeled:

\begin{itemize}
    \item \textbf{Potential User Persona:} The model often identifies a candidate audience (e.g., a student or someone interested in law''), seemingly calibrating its internal technicality accordingly.
    \item \textbf{Structural Planning:} We observe instances where the model pre-defines a structural schema (e.g., listing each right with support from legal frameworks''), potentially guiding the final response toward a more deliberate organizational logic.
\end{itemize}

To investigate whether these qualitative observations represent a broader trend, we quantify the frequency of these two specific patterns—\textit{User Needs Identification} and \textit{Structural Output Planning}—across the evaluation set.

As shown in Table \ref{tab:think_pattern}, there is a notable shift in the occurrence of these patterns. While vanilla GRPO identifies user needs in 54.78\% of cases, this frequency rises to 90.45\% in the START-trained model. Similarly, proactive structural planning is observed in 92.36\% of the traces. 

While these patterns alone may not fully account for the final performance gains, their systematic emergence suggests that START encourages the model to explore the observed modeling of meta-contexts. This emergence provides a plausible window into how independent thinking optimization might reshape the model's approach to general-domain instructions, moving beyond simple retrieval toward a more context-aware internal process.

\begin{table}[ht]
    \centering
    \small
    \caption{\textbf{Quantitative comparison of thinking patterns.} We report the frequency of meta-context dimensions (\textit{User Needs} and \textit{Structural Output}) identified within the thinking traces across the test set.}
    \label{tab:think_pattern}
    \begin{tabularx}{0.61\columnwidth}{l|ccc}
    \toprule
    \textbf{Pattern} & \textbf{GRPO} & \textbf{GRPO-MA} & \textbf{GRPO+\methodname} \\
    \midrule
    User Needs & 54.78\% & 61.78\% & 90.45\% \\
    Structural Output & 61.15\% & 75.99\% & 92.36\% \\
    \bottomrule
    \end{tabularx}
\end{table}
\section{Related Work}
\label{sec:related_work}

\subsection{Thinking Models and RLVR}

In recent years, the focus of Large Language Models (LLMs) research has shifted from simple next-token prediction toward the development of complex reasoning capabilities. Central to this evolution is the emergence of the "Thinking-Answer" paradigm, where models generate a Chain-of-Thought (CoT)\cite{cot, self_consistency} as an intermediate reasoning step before producing a final response.

The advent of Reinforcement Learning from Verifiable Rewards (RLVR)\cite{r1, deepseekv3} has significantly pushed the boundaries. Modern reasoning models, such as OpenAI o1 and DeepSeek-R1, have demonstrated that RL can substantially enhance reasoning performance in tasks with verifiable outcomes, such as mathematics, coding, and formal logic\cite{verify, gsm8k, math, human_eval}. In these domains, the environment provides objective and precise reward signals—such as the correctness of a mathematical result or the pass rate of code—allowing the model to evolve its reasoning strategies through large-scale self-play and trajectory search. However, current RLVR research remains largely confined to these rule-based, vertical domains. Whether the reasoning capabilities acquired through RLVR can seamlessly generalize to General Question Answering (GQA) tasks remains an open question\cite{math2tran}.



\subsection{LLMs in General QA}

The pursuit of human-level performance on General Question Answering (GQA) remains a central objective in the evolution of Large Language Models\cite{constitutionalai}. To measure progress in this area, the community has developed a suite of sophisticated benchmarks such as AlpacaEval 2.0\cite{alpaca_eval}, Arena-Hard\cite{arena_hard}, and WildBench\cite{wildbench}, which utilize LLM-as-a-judge frameworks\cite{llm_judge} to capture the nuances of human-like responses and stylistic alignment. Recently, the "thinking" paradigm has been integrated into these general-purpose applications. Frontier models, most notably OpenAI o1\cite{openai_o1} and DeepSeek-R1, have offered thinking mode for enhanced performance in everyday scenarios beyond reasoning.

Despite the industrial push for "thinking" models, scholarly investigation into the specific mechanics and efficacy of thinking processes for non-verifiable general tasks is still in its infancy. A prominent exception is ``Thinking LLMs''\cite{thinking_llm}, which introduced Thought Preference Optimization (TPO). This work demonstrates that models can be trained to think in general domains through an iterative search and optimization procedure without direct human thought supervision. However, there remains a critical gap in understanding whether RLVR naturally generalize to GQA.

\section{Conclusion}
\label{sec:conclusion}

In this work, we have investigated the untapped potential of internal thinking processes in GQA. By introducing a Cross-Generation evaluation framework, we revealed a significant disparity between domains: while thinking processes in reasoning tasks serve as a definitive performance multiplier, their efficacy in GQA is significantly lower. Moreover, we highlights that: unlike the transformative gains observed in reasoning-heavy tasks through RLVR, direct reinforcement learning on GQA tasks often fails to stimulate substantive thinking evolution. Base on the hypothesize that this failure stems from a ``loose coupling'' between thinking and answering in general tasks, we proposed \methodname. By isolating the thinking phase during the initial stages of training, \methodname~forces the model to focus on the cognitive quality of its reasoning manifold. Our experimental results across multiple benchmarks and algorithms consistently demonstrate the effectiveness of \methodname. Ultimately, we hope this exploration encourages further research into how internal thought processes can be more effectively cultivated to serve as a functional engine for broader AI applications.

\bibliography{reference}
\bibliographystyle{iclr2026_conference}

\newpage
\appendix

\section{Datasets and Hyperparameters}
\label{app:data}

In this section, we provide a detailed overview of the datasets utilized in our study and the specific configurations used for evaluation and training.

\subsection{MATH Dataset}

The MATH dataset\cite{math} is a widely recognized benchmark designed to evaluate the mathematical problem-solving capabilities of large language models across various subjects, including algebra, geometry, and calculus. Problems in this dataset are categorized into five difficulty levels. To provide a more rigorous assessment of the model’s "thinking" depth and avoid performance saturation, we focus specifically on the Level 5 subset, which contains the most challenging problems. We randomly sampled 2,000 problems for the training set and a separate, distinct set of 200 problems for the test set. To elicit structured reasoning traces, we add the following prompt to the original question:

\begin{tcolorbox}[
    colback=gray!5!white, 
    colframe=gray!75!black, 
    title=Prompt Template, 
    fonttitle=\bfseries,
    boxrule=0.5pt,
    arc=2pt
]
\texttt{Let's think step by step and output the final answer within \textbackslash boxed\{\}.}
\end{tcolorbox}

For verifiable reward calculation, we employ the answer extraction and normalization functions provided by the verl library\cite{verl} to ensure consistent and accurate scoring against ground-truth solutions.

\subsection{AlpacaEval 2.0 Benchmark}

AlpacaEval 2.0\cite{alpaca_eval} is an automatic evaluator for general question answering, designed to simulate human preferences across a diverse set of 805 prompts. These prompts span various real-world categories, including creative writing, coding assistance, and general information seeking. As recommended, We report the Length-controlled (LC) Win Rate as our main performance indicator. This metric is specifically designed to mitigate the "length bias"—a common issue where evaluators favor longer responses regardless of actual quality—by adjusting the win rate based on the relative length of the model's output versus the reference.

To balance computational efficiency with evaluation accuracy, we utilize the \textit{alpaca\_eval\_vllm\_llama3\_70b\_fn} as the evaluator. This evaluator is highly ranked in terms of human agreement while offering the distinct advantage of local deployment via the vLLM framework\cite{vllm}.

During the evaluation of the 805 samples, we observed that a negligible number of instances—typically 2 to 3 samples per run—encountered parsing errors or failed to produce a readable score from the evaluator. Given the extremely low frequency of these occurrences (less than 0.4\% of the total set), they do not statistically impact the final calculated win rate.

\subsection{ExpertQA Dataset}

ExpertQA\cite{expertqa} is a high-quality benchmark designed to evaluate the performance of large language models on complex, domain-specific questions verified by experts. The dataset spans a wide array of academic and professional fields, requiring both factual accuracy and nuanced reasoning.

We utilize the official main set \textit{data/r2\_compiled\_anon.jsonl} file. This version contains the anonymized, compiled data from the second round of expert reviews, ensuring a high standard of reference quality and expert-level alignment. To maintain a consistent evaluation framework while ensuring the model is exposed to a diverse set of expert queries during training, 90\% of the samples were randomly selected to serve as the training corpus for optimizing the thinking traces and responses. The remaining 10\% of the data was reserved for testing.

\subsection{UltraFeedback Dataset}

UltraFeedback\cite{ultrafeedback} is a large-scale, high-quality preference dataset designed to align language models with human intentions. It provides a diverse range of prompts and multifaceted feedback across dimensions such as instruction-following, truthfulness, and honesty.

To maintain training stability and efficiency, we utilize a curated version of the dataset ``trl-lib/ultrafeedback-prompt'' from Hugging Face. This version specifically excludes samples with excessively long prompts, ensuring that the context window is focused on the interaction between the thinking trace and the final answer rather than processing outlier input lengths.

We randomly sampled 2,000 instances from the training split to serve as the training corpus and 200 instances from the test split to measure the final performance and the effectiveness of the thinking evolution on unseen prompts.

\subsection{\textbf{Sampling Parameters}}

In our experiments, we use large language models of three series. For both training and testing phases, we strictly adhere to the sampling parameters recommended as best practices for each model family to ensure optimal performance and representative behavior. Table \ref{tab:sampling_params} summarizes the specific configurations used for the thinking and answering generation.

\begin{table}[h]
    \centering
    \caption{Sampling parameters used across different model families.}
    \label{tab:sampling_params}
    \begin{tabularx}{0.7\linewidth}{l|ccc}
    \toprule
    \textbf{Parameter} & \textbf{Qwen3} & \textbf{DeepSeek-R1-Distill} & \textbf{Hunyuan} \\
    \midrule
    Temperature & 0.6 & 0.6 & 0.7 \\
    Top-$p$ & 0.95 & 0.95 & 0.8 \\
    Top-$k$ & 20 & 20 & 20 \\
    Min-$p$ & 0 & 0 & 0 \\
    Do Sample & True & True & True \\
    Repetition Penalty & 1.0 & 1.0 & 1.05 \\
    \bottomrule
    \end{tabularx}
\end{table}

\subsection{Reward Model Configuration}

For all experiments involving continuous reward signals in this study, we utilize ArmoRM-Llama3-8B-v0.1 as the primary reward model. ArmoRM\cite{ArmoRM} is a state-of-the-art reward model based on the Llama-3-8B architecture, designed to align large language models with human preferences. Unlike traditional reward models that provide a single scalar output, ArmoRM is a multi-objective reward model trained on a diverse set of preference datasets. It is capable of evaluating multiple dimensions of response quality, such as helpfulness, truthfulness, and safety. In our implementation, we use the overall score provided by the model. This score represents a holistic assessment of the response quality by aggregating the various attribute-specific rewards into a single scalar value.

\section{Reproduction Details and RL Environments}

To ensure the reproducibility of our findings, we provide the following details regarding the reinforcement learning framework and training environment.

All reinforcement learning training and evaluation pipelines are implemented using the verl library\cite{verl}. We have included the complete training scripts, evaluation code, and configuration files in the Supplementary Material. The full codebase will be open-sourced upon the acceptance of this paper to facilitate further research into thinking evolution.

\subsection{env}

To maintain consistency and avoid version-related performance variations, all experiments were conducted within a standardized containerized environment. We utilize the following Docker image provided by the verl team: ``verlai/verl:app-verl0.5-transformers4.55.4-vllm0.10.0-mcore0.13.0-te2.2''. This environment includes optimized versions of the Transformers library, vLLM for high-throughput inference, and fsdp for scalable training.

\subsection{GRPO}

The specific hyperparameter configurations for our GRPO experiments are detailed in Table \ref{tab:grpo_params}.

\begin{table}[htbp]
\centering
\caption{GRPO training hyperparameters and configuration details}
\label{tab:grpo_params}
\begin{tabular}{lll}
\toprule
\textbf{Category} & \textbf{Parameter} & \textbf{Value} \\
\midrule
\multicolumn{3}{l}{\textit{Training Setup}} \\
& Save frequency & per epoch \\
& Test/evaluation frequency & per epoch \\
& Training batch size & 64 \\
& Max prompt length & 256 tokens \\
& Max response length & 3,328 tokens \\
\midrule
\multicolumn{3}{l}{\textit{Model Configuration}} \\
& Model dtype & \texttt{bfloat16} \\
& Gradient checkpointing & \texttt{True} \\
& Remove padding & \texttt{True} \\
& Max think tokens & 2,047 \\
& Max answer tokens & 1,280 \\
\midrule
\multicolumn{3}{l}{\textit{Optimization}} \\
& Learning rate & $1 \times 10^{-5}$ \\
& PPO mini-batch size & 16 \\
& PPO micro-batch size per GPU & 2 \\
\midrule
\multicolumn{3}{l}{\textit{KL Regularization}} \\
& KL loss enabled & \texttt{True} \\
& KL coefficient & 0.001 \\
& KL type & \texttt{low\_var\_kl} \\
& KL in reward & \texttt{False} \\
\midrule
\multicolumn{3}{l}{\textit{Entropy \& Exploration}} \\
& Entropy coefficient & 0.0 \\
& Rollout samples per prompt ($n$) & 6 \\
& Validation sampling & \texttt{True} \\
\midrule
\multicolumn{3}{l}{\textit{Rollout Engine}} \\
& Log prob micro-batch size & 16 \\
\midrule
\multicolumn{3}{l}{\textit{Reward Configuration}} \\
& Reward estimator & GRPO \\
\bottomrule
\end{tabular}
\end{table}

\subsection{DAPO}

The implementation of DAPO is based on the standard recipes provided within the verl library, and the specific hyperparameter configurations are detailed in Table \ref{tab:dapo_params}.

\begin{table}[htbp]
\centering
\caption{DAPO training hyperparameters and configuration details}
\label{tab:dapo_params}
\begin{tabular}{lll}
\toprule
\textbf{Category} & \textbf{Parameter} & \textbf{Value} \\
\midrule
\multicolumn{3}{l}{\textit{Training Setup}} \\
& Save frequency & per epoch \\
& Test/evaluation frequency & per epoch \\
& Training batch size & 64 \\
& Generation batch size & 64 \\
& Max prompt length & 256 tokens \\
& Max response length & 3,328 tokens \\
\midrule
\multicolumn{3}{l}{\textit{Model Configuration}} \\
& Model dtype & \texttt{bfloat16} \\
& Gradient checkpointing & \texttt{True} \\
& Remove padding & \texttt{True} \\
& Max think tokens & 2,047 \\
& Max answer tokens & 1,280 \\
\midrule
\multicolumn{3}{l}{\textit{Optimization}} \\
& Learning rate & $1 \times 10^{-5}$ \\
& LR warmup steps & 10 \\
& Weight decay & 0.1 \\
& Gradient clipping norm & 1.0 \\
& Loss aggregation mode & \texttt{token-mean} \\
& PPO mini-batch size & 16 \\
& PPO micro-batch size per GPU & 2 \\
& Dynamic batch size & \texttt{True} \\
\midrule
\multicolumn{3}{l}{\textit{KL Regularization}} \\
& KL loss enabled & \texttt{False} \\
& KL in reward & \texttt{False} \\
\midrule
\multicolumn{3}{l}{\textit{DAPO-Specific Policy Clipping}} \\
& Clip ratio lower bound & 0.20 \\
& Clip ratio upper bound & 0.28 \\
& Clip ratio curvature ($c$) & 10.0 \\
\midrule
\multicolumn{3}{l}{\textit{Entropy \& Exploration}} \\
& Entropy coefficient & 0.0 \\
& Rollout samples per prompt ($n$) & 6 \\
& Validation sampling & \texttt{True} \\
\midrule
\multicolumn{3}{l}{\textit{Rollout Engine (vLLM)}} \\
& Log prob micro-batch size & 16 \\
& Dynamic batch size (rollout) & \texttt{True} \\
\midrule
\multicolumn{3}{l}{\textit{Group Filtering (DAPO)}} \\
& Enable filter groups & \texttt{False} \\
& Max generation batches & 10 \\
& Filtering metric & \texttt{acc} \\
\midrule
\multicolumn{3}{l}{\textit{Overlong Buffer Reward}} \\
& Enable overlong buffer & \texttt{True} \\
& Buffer length & 2,560 tokens \\
& Penalty factor & 1.0 \\
\midrule
\multicolumn{3}{l}{\textit{Reward Configuration}} \\
& Reward estimator & GRPO \\
& Reward manager & \texttt{dapo} \\
\bottomrule
\end{tabular}
\end{table}

\subsection{GRPO-MA}

For GRPO-MA, We modified the sampling engine within the verl library to support a two-tier generation strategy, generate multiple independent answers for each thinking trace and calculate advantage independently.

To maintain a fair comparison with standard GRPO in terms of total sampled tokens, we only adjusted the group composition. While the base RL hyperparameters (learning rate, KL coefficient, etc.) remain identical to those in Table \ref{tab:grpo_params}, the sampling structure is: 2 thinking traces per prompt, 3 answers per thinking trace, $2*3=6$ total samples per prompt.

\section{Analysis of Thinking-Answer Coupling across Domains}
\label{app:coupling_analysis}

To explore the coupling disparities between different task categories, we conduct exploratory experiments on the MATH and ExpertQA datasets, representing reasoning-intensive and general question answering domains, respectively. The objective of these experiments is to preliminary explore the coupling characteristics between thinking traces and final responses across different task types.

\subsection{sampling method}

To formalize the observation of coupling disparities, we define a nested sampling and scoring framework. For a given instruction $x$, we first sample a set of $m$ distinct thinking traces $\mathcal{T} = \{T_j\}_{j=1}^m$. Subsequently, for each fixed thinking trace $T_j$, we sample $n$ independent response sequences $\{A_{j,k}\}_{k=1}^n$. For each complete generation path $s_{j,k} = [T_j, A_{j,k}]$, we obtain a quality score $R_{j,k}$. Specifically, for the MATH dataset, $R_{j,k}$ is a binary verifiable reward based on the correctness of the final answer; for ExpertQA, $R_{j,k}$ is a continuous scalar provided by a reward model.

Follow GRPO-MA, we define the Thinking Score as the expected score of all possible answers derived from a specific thought. This serves as a measure of the "inherent value" of a thinking trace, independent of the subsequent answering stochasticity. Based on the nested sampling framework, for each thinking trace $T_j$, the thinking score $S_j$ is defined as:
\begin{equation}
S_j = \mathbb{E}_{A \sim \pi(\cdot|x, T_j)} [r(x, T_j, A)] \approx \frac{1}{n}\sum_{k=1}^{n} R_{j,k}
\end{equation}

This score represents the potential of the thinking trace $T_j$ to steer the model toward a correct or high-quality response. In our experiments, we set $m=8$ and $n=8$, resulting in 64 complete trajectories for each instruction $x$ to ensure a statistically robust estimation.

\subsection{Strong Coupling in Reasoning Tasks}

To provide a concrete measure of coupling in the reasoning domain, we first analyze the outcome consistency for each fixed thinking trace on the MATH dataset. Since MATH provides verifiable binary rewards (correct or incorrect), we define the \textbf{Minority Outcome Ratio} $\gamma$ to quantify the stochasticity of the answering phase relative to the thinking process.

For a given instruction $x$ and a specific thinking trace $T_j$, let $n_{j,1}$ and $n_{j,0}$ denote the number of correct and incorrect answers among the $n$ samples, respectively. The ratio $\gamma$ is calculated as:

\begin{equation}
\gamma = \mathbb{E}_{x, T_j} \left[ \frac{\min(n_{j,1}, n_{j,0})}{n} \right]
\end{equation}

where the expectation is taken over all sampled thinking traces across the dataset. Our empirical analysis reveals that $\gamma$ is merely \textbf{0.0469}.

This strikingly low value aligns with the intuitive nature of mathematical reasoning: the correctness of a solution is almost entirely determined by the logical integrity of the preceding thinking process. In this ``strong coupling'' scenario, the answering phase acts as a nearly deterministic transducer; once a correct logical path is established in $T$, the probability of generating an incorrect final answer $A$ is negligible. Conversely, a logical error in $T$ typically precludes a correct $A$. Thus, the minority outcome ratio of $0.0469$ is a direct reflection of the strong coupling in reasoning tasks, where the quality of the thinking trace leaves little room for stochastic fluctuation in the final answer.

\subsection{Loose Coupling in General QA}

Unlike reasoning tasks with binary outcomes, general-purpose tasks like ExpertQA involve continuous quality scores. To analyze the coupling in this domain, we utilize the metrics of \textbf{Answering Fluctuation} ($\sigma_{\text{answer}}$) and \textbf{Thinking Fluctuation} ($\sigma_{\text{thinking}}$) as defined below. For each instruction $x$, based on the $m \times n$ sampled trajectories and their corresponding reward scores $R_{j,k}$, we compute:

\begin{itemize}
    \item \textbf{Answering Fluctuation ($\sigma_{\text{answer}}$)}: This represents the average degree of quality variation that occurs during the response generation phase for a fixed thought.:
        \begin{equation}
            \sigma_{\text{answer}}=\text{mean}_{j}\left(\text{std}_{k}(R_{j,k})\right)
        \end{equation}
    where $S_j = \frac{1}{n} \sum_{k=1}^n R_{j,k}$ is the thinking score for the $j$-th trace.
    \item \textbf{Thinking Fluctuation ($\sigma_{\text{thinking}}$)}: This represents the variation in potential quality across different thinking paths. It measures how much the "direction of thought" actually influences the expected reward:
        \begin{equation}
            \sigma_{\text{thinking}}=\text{std}_{j} (S_j)
        \end{equation}
    where $S_j = \frac{1}{n} \sum_{k=1}^n R_{j,k}$ is the thinking score defined above.
\end{itemize}

We then define the \textbf{Coupling Ratio} $\rho = \sigma_{\text{thinking}} / \sigma_{\text{answer}}$. A ratio $\rho < 1.0$ implies that the quality signal from thinking is frequently overshadowed by the fluctuations in the answering phase, indicating a state of ``loose coupling''.

\begin{figure}[ht]
    \centering
    \includegraphics[width=0.6\linewidth]{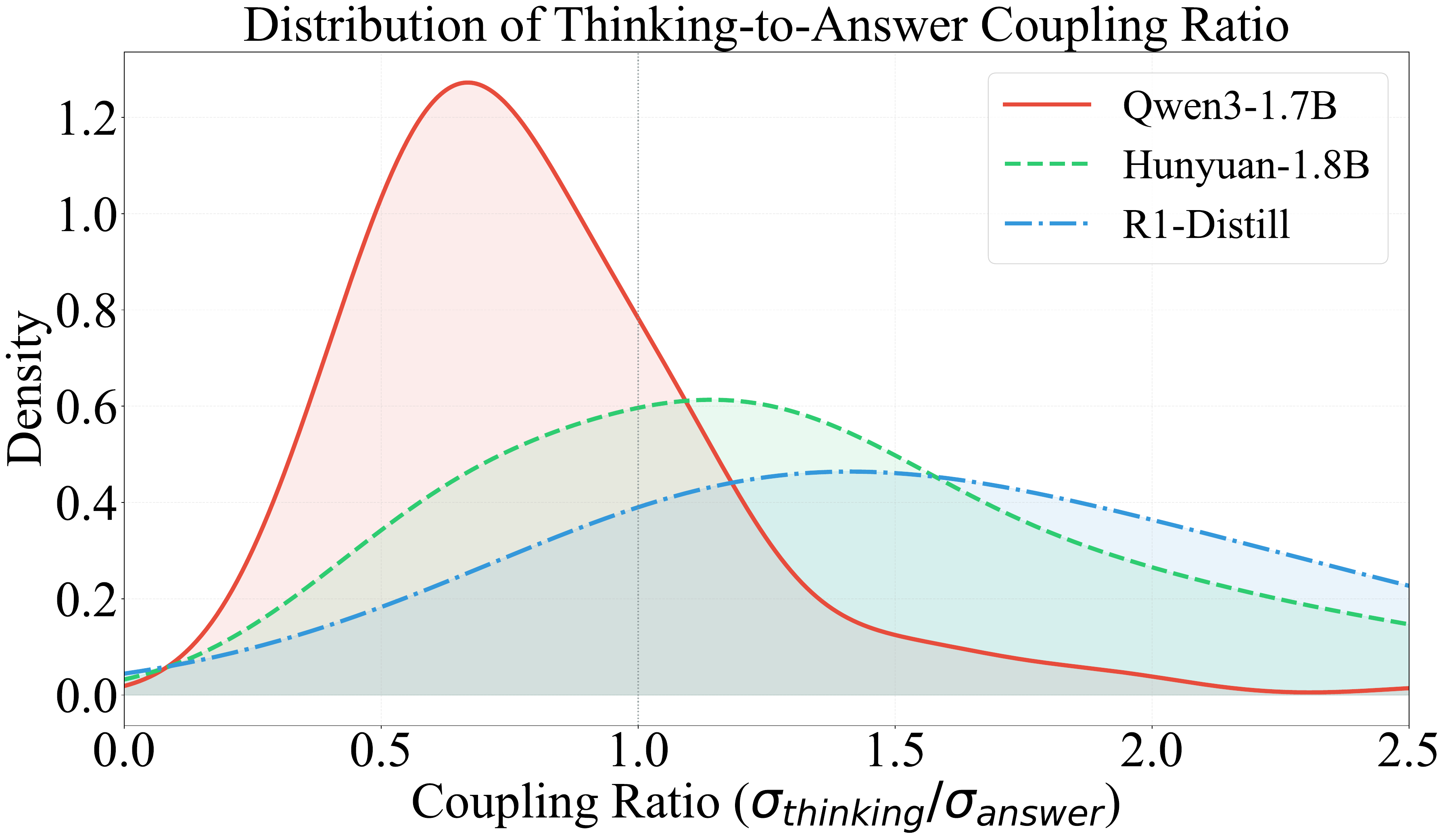}
    \caption{\textbf{Distribution of thinking-to-answer Coupling Ratio on ExpertQA.} The red curve (Qwen3-1.7B) exhibits a sharp peak at $Ratio \approx 0.65$.}
    \label{fig:ratio_dis}
\end{figure}

Based on these metrics, we visualize the distribution of the Coupling Ratio $\rho$ across the ExpertQA dataset in Figure \ref{fig:ratio_dis}. These empirical results provide a preliminary exploration of the loose coupling inherent in general tasks, revealing a starkly different dynamic compared to the consistency observed in MATH.

In this general-purpose domain, the quality fluctuation during the answering phase is remarkably high—frequently comparable to or even exceeding the fluctuation attributed to the thinking process. For the majority of samples across the evaluated models, the density peaks remain predominantly below or near the critical $\rho = 1.0$ line. This indicates that, unlike the deterministic nature of reasoning tasks, the thinking trace in GQA does not consistently dictate the ``result''. Instead, the final output quality is heavily influenced by stylistic, structural, or linguistic variances in the answering phase, even when the underlying reasoning remains the same.

While models such as R1-Distill show a relative right-shift compared to smaller models like Qwen3-1.7B, it is important to note that none of the models achieve a state of universal strong coupling in the general domain. Even for stronger models, a significant portion of the distribution resides in the loose coupling zone ($\rho < 1.0$). This demonstrates that the presence of loose coupling is a pervasive characteristic of GQA tasks across models of varying scales.

\section{Example for Meta-Context}

\definecolor{basecolor}{RGB}{100, 100, 100}
\definecolor{refcolor}{RGB}{200, 150, 0}
\definecolor{startcolor}{RGB}{0, 100, 0}

\begin{tcolorbox}[
    enhanced,
    title=\textbf{Evolution of Thinking Patterns},
    colback=white,
    colframe=black!75,
    fonttitle=\small\bfseries,
    arc=2mm
]
    \begin{tcolorbox}[colback=blue!5, colframe=blue!20, left=2mm, right=2mm, top=1mm, bottom=1mm]
        \scriptsize \textbf{Instruction:} What are the fundamental rights of the accused?
    \end{tcolorbox}
    \vspace{1mm}
    \begin{tcolorbox}[colback=gray!5, colframe=basecolor!50, title=\footnotesize \textbf{Base Model (Qwen3-1.7B)}, left=1mm, right=1mm, top=1mm, bottom=1mm]
        \scriptsize
        ``...I need to figure out... I think I have a list, but I need to organize them properly...'' \\
    \end{tcolorbox}
    \vspace{-2mm}
    \begin{tcolorbox}[colback=orange!5, colframe=refcolor!50, title=\footnotesize \textbf{Vanilla GRPO}, left=1mm, right=1mm, top=1mm, bottom=1mm]
        \scriptsize
        ``...First, the accused has the right to a fair trial... I need to make sure the answer is comprehensive but not too technical... so clarity is important. Let me organize these points in a logical order.'' \\
    \end{tcolorbox}
    \vspace{-2mm}
    \begin{tcolorbox}[colback=green!5, colframe=startcolor!70, title=\footnotesize \textbf{GRPO+\methodname}, left=1mm, right=1mm, top=1mm, bottom=1mm]
        \scriptsize
        ``...start by recalling key principles... \textbf{[Structural Planning] I should structure the answer clearly, listing each right with support from legal frameworks...} I need to make sure the language is accessible and not too technical, \textbf{[User Needs] so the user, whether they're a student or someone interested in law, can understand the points without getting lost in jargon}. Finally, a concluding statement that reinforces the value of these rights...'' \\
        \textcolor{startcolor}{\rule{\linewidth}{0.2pt}} \\
        \textbf{Improvements:} Demonstrates meta-contexts like audience needs and structural design.
    \end{tcolorbox}
\end{tcolorbox}

\section{Prompts for Meta-Context Identification}

\begin{tcolorbox}[
    colback=gray!5!white, 
    colframe=gray!75!black, 
    title=Prompt for User Intent and Identity Speculation Detection, 
    fonttitle=\bfseries,
    boxrule=0.5pt,
    arc=2pt
]
\texttt{
Please evaluate whether the provided 'thinking' content from a large language model contains speculation about the user's identity, role, persona, or hidden intent and psychology.\\
Definitions: *User Identity or Intent Speculation involves:\\
The model explicitly attempts to infer who the user is, their professional background, their level of expertise, or their specific role (e.g., "The user seems to be a student," "The user is likely a developer," "Assuming the user is a non-expert").\\
The model analyzes the prompt to guess the user's hidden intent, underlying motivation, or psychological state (e.g., "The user might be trying to test my safety filters," "The user seems frustrated," or "The underlying goal is likely to bypass a restriction").\\
To do so, I will give you the instructions (prompts) given to the model, and the thinking process of the model.\\
All inputs should be Python dictionaries.\\
Here is the prompt: \{ "instruction": "\{\{instruction\}\}", \}\\
Here is the thinking process of the model: \{ "thinking": "\{\{context\}\}" \}\\
Requirements for your reply:\\
Output ONLY the character '1' (if user identity, intent, or psychological speculation is present) or '0' (if no such speculation is detected).\\
DO NOT include any other text, explanation, punctuation, or code blocks.\\
The output must be a single integer.
}
\end{tcolorbox}

\begin{tcolorbox}[
    colback=gray!5!white, 
    colframe=gray!75!black, 
    title=Prompt for Formatting and Stylistic Planning Detection, 
    fonttitle=\bfseries,
    boxrule=0.5pt,
    arc=2pt
]
\texttt{
Please evaluate whether the provided 'thinking' content from a large language model contains any mention of how to structure, style, or present the final answer.\\
Definitions: Format or Tone Specification (Output 1 if any of these exist):\\
Structural/Planning Intent: Any mention of how the response should be organized or what it should contain (e.g., "I should cover...", "include factors", "mention X and Y", "list the steps", "start by...", "give examples").\\
Tone/Style/Vibe: Any mention of the desired feel or level of detail (e.g., "be comprehensive", "be confident", "keep it simple", "make it professional", "avoid being too technical", "be concise").\\
Target Goal: Any statement starting with "The answer should be..." or "I need to make sure the response is..." followed by an adjective (e.g., "comprehensive", "clear", "helpful").\\
Input Data:\\
Prompt: \{\{instruction\}\}\\
Thinking Process: \{\{context\}\}\\
Requirements for your reply:\\
Output ONLY the character '1' (if ANY mention is found) or '0' (if none).\\
DO NOT include any other text.\\
}
\end{tcolorbox}

\end{document}